\documentclass[conference]{IEEEtran}
\IEEEoverridecommandlockouts
\usepackage{cite}
\usepackage{amsmath,amssymb,amsfonts}
\usepackage{algorithmic}
\usepackage{graphicx}
\usepackage{textcomp}
\usepackage[table]{xcolor}
\usepackage[inline]{enumitem}

\def\BibTeX{{\rm B\kern-.05em{\sc i\kern-.025em b}\kern-.08em
    T\kern-.1667em\lower.7ex\hbox{E}\kern-.125emX}}
    
\newcommand{\droso}{\textsc{MediDelivery}}

\usepackage{caption}
\usepackage{subcaption}
\usepackage{array}
\usepackage{multirow}
\usepackage{tablefootnote}

\definecolor{lightgray}{gray}{0.9}

\begin{document}

\title{Certifying Emergency Landing for Safe Urban UAV \\
\thanks{Our work has benefited from the AI Interdisciplinary Institute ANITI. ANITI is funded by the French "Investing for the Future – PIA3" program under the Grant agreement n°ANR-19-PI3A-0004.}
}

\author{\IEEEauthorblockN{Joris Guerin}
\IEEEauthorblockA{\textit{Université Toulouse, LAAS-CNRS, ONERA} \\
Toulouse, France \\
joris.guerin@laas.fr}
\and
\IEEEauthorblockN{Kevin Delmas}
\IEEEauthorblockA{\textit{ONERA} \\
Toulouse, France \\
kevin.delmas@onera.fr}
\and
\IEEEauthorblockN{Jérémie Guiochet}
\IEEEauthorblockA{\textit{Université Toulouse, LAAS-CNRS} \\
Toulouse, France \\
jeremie.guiochet@laas.fr}
}

\maketitle

\begin{abstract}
Unmanned Aerial Vehicles (UAVs) have the potential to be used for many applications in urban environments. However, allowing UAVs to fly above densely populated areas raises concerns regarding safety. One of the main safety issues is the possibility for a failure to cause the loss of navigation capabilities, which can result in the UAV falling/landing in hazardous areas such as busy roads, where it can cause fatal accidents. Current standards, such as the SORA published in 2019, do not consider applicable mitigation techniques to handle this kind of hazardous situations. Consequently, certifying UAV urban operations implies to demonstrate very high levels of integrity, which results in prohibitive development costs. To address this issue, this paper explores the concept of Emergency Landing (EL). A safety analysis is conducted on an urban UAV case study, and requirements are proposed to enable the integration of EL as an acceptable mitigation mean in the SORA. Based on these requirements, an EL implementation was developed, together with a runtime monitoring architecture to enhance confidence in the system. Preliminary qualitative results are presented and the monitor seem to be able to detect errors of the EL system effectively.
\end{abstract}


\section{Introduction}\label{sec:intro}
%Paragraph 1: Drones in urban environment beneficial but dangerous
The range of usage of Unmanned Aerial Vehicles (UAV) is growing rapidly and includes many application domains with positive impact for society such as agriculture, construction, structures visual inspection, live events filming, or delivery of medical equipment~\cite{drones_urban_environment, medical_delivery}. However, their use in urban environment is still limited by technical and operational challenges related to societal acceptance and safety. As explained in~\cite{drones_urban_environment}, the primary risks for UAVs in urban environments are collision with other flying systems operating at low altitudes (air risk) and impact with third party below and adjacent to the area of operation (ground risk). 

%Paragraph 2: In europe, the SORA was introduced a few years ago as a norm to certify specific UAV urban operations. However, with the current state of the SORA, it is near impossible to certify urban operations.
In most developed countries, an operation conducted by UAV can only be authorized when it is approved by a regulatory administration (e.g., EASA~\cite{EASA_opinion}, FAA~\cite{FAA_rules}). In particular, the Joint Authorities for Rulemaking of Unmanned Systems (JARUS) recently released a document called Specific Operations Risk Assessment (SORA)~\cite{SORA} to present official guidelines for the development of UAV operations that can be authorized by competent authorities. The SORA presents a detailed procedure to evaluate the risks of a specific UAV operation, as well as potential actions to mitigate these risks and precise conditions under which the operation can be authorized. The guidelines for certification are usually to keep the UAV far from people, properties and surrounding aircraft traffic to reduce risk if a critical failure occurs. For urban environments, this separation cannot be achieved and intensive risk analysis must be conducted to demonstrate that the UAV flight safety is acceptable for such high risk environment. In practice, this process is very demanding and most civil UAV urban operations cannot be authorized.

%Paragraph 3: EL exists and seems a good strategy to enhance safety of urban operations. However, the field of EL lacks a unified definition of what are the objectives of a successful EL algorithm. In addition, EL is not included as a valid mean of mitigation in the SORA.
Regarding the ground risk, several Emergency Landing (EL) methods have been proposed to reduce the risk to people, infrastructures and the UAV itself~\cite{risk_from_database, edges_from_high_altitude_1, depth_and_flat}. As a general definition, EL aims at leveraging available information (on-board sensors, external databases) to select an area where the UAV can land safely, and to reach this landing zone in a potentially degraded control mode. Nevertheless, to this day, despite the noteworthy advances in this field of research, the precise requirements for EL are still unclear and no consensus is emerging about safety objectives. Because of this lack of well defined requirements, EL is still not considered as a valid mitigation to lower the ground risk in the SORA, and thus cannot be leveraged to ease the certification process of UAV urban operations. %Hence, the first goal of this work is to clearly formulate what should be expected from an EL module.
%As EL is undoubtedly intended to reduce the risk to harm people in case of failures, it should be considered as a mean of mitigation. However, it is still not the case in the current version of the SORA, and EL cannot be leveraged to ease the certification process. This is probably due to the lack of maturity of current EL methods, which still cannot be properly evaluated because of the lack of well defined requirements. In this way, the second objective of this paper is to understand if EL can be considered a relevant safety function, and under which conditions. In particular, we want to know if EL could be integrated within the SORA process and how this should be done.

%Paragraph 4: Contributions
The first contribution of this work consists in studying a specific use case called \droso, to show that the current state of the SORA has limitations when applied to urban operations. The second contribution is to propose a set of safety requirements for EL, which could be used within the SORA to evaluate an EL module as a valid mitigation, thus easing the certification process of urban operations. Finally, we show that successful EL implementations must rely on complex computer vision functions, often based on deep learning, on which it is difficult to obtain safety guarantees. Hence, to comply with the proposed mitigation requirements, new advanced runtime monitoring techniques must be developed to ensure safety. The final contribution of this work is to propose such an implementation of EL, using semantic segmentation to identify landing zones and Bayesian neural network theory to monitor the predictions of the EL system at runtime. Primary qualitative results on the proposed solution are presented and show that our approach is promising and is worth further investigations.

\section{Background}\label{sec:background}

%This section presents useful previous work regarding safety and certification of UAV operations as well as literature related to EL.

\subsection{Certifying UAV operations}\label{sec:sora_overview}
As for manned aviation, a European regulation (namely the EU 2019/947 and 2019/945~\cite{EASA_reg}) has been published by the European Union Aviation Agency (EASA) for UAVs. This regulation provides an implementation of rules and procedures to which applicant must demonstrate its compliance to obtain a flight authorization. Due to the variety of drones (ranging from toys to aircraft-like drones) and operations (\textit{e.g.} infrastructure monitoring v.s. crowd surveillance) the severity of the possible outcomes of a UAV failure may be drastically different. This is why the regulation considers three categories of UAV operations:
\begin{description}
\item[Open] light UAVs operating at a safe distance from people. These operations are not subject to any prior flight authorization due to minor severity of their failure.
\item[Specific] heavier UAVs operating nearby populated areas. These operations are subject to prior flight authorization and conduct of a Specific Operation Risk Assessment (SORA)~\cite{SORA}.
\item[Certified] heaviest UAVs operating over assemblies of people  or transporting people or dangerous goods. These operations are subject to manned aircraft like certification rules.
\end{description}

An important part of operations involving medium endurance drone, typically used for short-range delivery, falls into the \textit{specific} category. The certification objectives for these operations are identified by applying the SORA. This document can be seen as a tailoring guide used by the applicant to identify:
\begin{enumerate}
    \item The level of on-ground and mid-air collision risks of the operation, called ground risk class (denoted GRC) ranging from 1 (low risk) to 8 (high risk) and air risk class (denoted ARC) ranging from A (low risk) to D (high risk). The computation of the intrinsic GRC and initial ARC are based on the UAV specifications, the operational scenario and the characteristics of the airspace.
    \item The suitable mitigation means and an estimation of their risk reduction. The implementation of appropriate mitigation strategies can be used to decrease the GRC and ARC. For GRC, three mitigation types are identified by the SORA: 
    \begin{itemize}
        \item M1 - \textit{Strategic mitigation}: intends to reduce the number of people at risk on the ground by ensuring that the UAV remains far from high risk areas. It is achieved through preliminary studies to show that few people are present in a predefined ground safety buffer throughout the trajectory.
        \item M2 - \textit{Reduction of ground impact effects}: intends to reduce the effect of the UAV impact dynamics (i.e., area, energy, impulse, transfer energy, etc.). One example would be to open a parachute.
        \item M3 - \textit{Emergency Response Plan (ERP)}: should be defined in the event of loss of control of the operation. %These are emergency situations where the operation is in an unrecoverable state and in which the outcome of the situation highly relies on providence, could not be handled by a contingency procedure, and there is grave and imminent danger of fatalities.
    \end{itemize}
    \item The final level of safety required for the operation with respect to the final GRC/ARC. In the SORA, this level is called the Specific Assurance and Integrity Level (SAIL).
    \item The applicable operational safety requirements (called OSOs) and the expected level of compliance demonstration (called Robustness) commensurate to the SAIL.
\end{enumerate}

With the current version of the SORA, it is very hard to certify UAV urban operations, mostly because the proposed mitigation strategies do not apply in densely populated environments (see Section~\ref{sec:sora}). For such highly complex environments, new active and intelligent mitigation means are required. A promising approach to reduce the number of people at risk on the ground is to select actively appropriate areas for landing, this kind of methods are called Emergency Landing (EL).

\subsection{Emergency Landing (EL)}\label{sec:literatureEL}

Several recent research have studied the problem of EL, most of which focus specifically on Landing Zone Selection (LZS). The existing methods to LZS can be divided into three main categories:

\subsubsection{LZS from public databases} 
Interesting works have proposed to leverage knowledge from both static and dynamic external databases to select EL sites without infrastructures and with few people at risk. The model proposed in~\cite{parachute_from_database} uses publicly available maps to minimize the risk to humans (far from buildings), to properties (far from power lines and transportation ways) as well as to the UAV itself (terrain slope and roughness, water). In~\cite{risk_from_database}, the risk associated to EL is refined using dynamic cellphone usage data and takes the time of the day into consideration.

\subsubsection{LZS from on-board camera at high altitude}
The second kind of approaches to LZS consists in using images from on-board camera at high altitude. The method presented in~\cite{edges_from_high_altitude_2}, applies a Canny edge detector on images recorded at high altitude and selects areas with lower concentration of edges for landing. This approach is improved in~\cite{edges_svm} by applying a Support Vector Machine (SVM) to classify the selected landing zones into building, bitumen, trees, grass or water. %The family of methods based on edge detection is enhanced in~\cite{edge_database_vision}, where referenced infrastructures from public databases are used to refine the LZS algorithm. 
In~\cite{depth_and_flat}, a Convolutional Neural Network (CNN) is trained to estimate depth from RGB images while simultaneously selecting a flat surface for safe landing. In practice, the LZ selected often ends up being flat roof-tops, roads or grass areas. %Similarly, the work conducted in~\cite{rooftop_landing} consists in identifying rooftops areas adapted for EL. 
Another common practice for recognizing different landing surfaces is to split the entire image into small tiles, which are classified into different categories. In~\cite{lstm_tile_classif}, tiles are classified into one of water, trees, grass, bitumen or building categories, while in~\cite{manual_labeling_high_altitude}, small image patches are manually labeled as safe, not recommended or unsafe and a CNN is trained to predict the suitability of regions in the image for landing.

\subsubsection{LZS from on-board camera at low altitude} 
Finally, the last family of methods uses images recorded on-board at low altitude. In~\cite{gabor_svm_surface_classif}, a method using a SVM classifier on Gabor features is presented to distinguish between grass/soil, tree and inland water from UAV images. Regarding LZS, the authors claim that grass should be preferred for landing to avoid damaging or losing the UAV. %A method to classify the type of terrain is also proposed in~\cite{terrain_classif_low_altitude}, to distinguish between water, vegetation, asphalt and sand from close-to-ground images. 
In~\cite{low_altitude_flat}, information from lidar and binocular camera is fused to identify flat and safe ground areas for landing. Similarly, the method proposed in~\cite{landmark_and_obstacle}, simultaneously detects a landmark and evaluate its eligibility for landing if there is no obstacles. As the detections from these two papers are conducted from low altitude, they can evaluate the real-time availability of landing spots, and select the right time for landing. Finally, the algorithm proposed in~\cite{low_terrain_classif} evaluates terrain flatness, steepness, depth accuracy and energy consumption to select a landing spot from low altitude.

\subsubsection{Limitations of current approaches} 
%On the one hand, using external databases allows the autonomous UAV to roughly select landing areas which are potentially out of sight. However, relying only on databases - even dynamic ones - cannot account for the ongoing real-time situation encountered by the UAV, and should be complemented by local LZS using on-board sensors. On the other, by using images collected at run-time, the algorithm has information regarding the current situation to decide precisely where to land but has a more limited area of action. When the altitude for LZS decreases, the certainty that the landing conditions will not change increases, but the chances of finding a suitable area decrease as the range of view becomes smaller.
This literature review reveals the lack of unified goals definition for EL. Although the general objective is to avoid hurting people, infrastructures and the UAV, the precise way to achieve it differs among different works. Indeed, while some studies consider flat areas, such as roads, as safe for landing~\cite{depth_and_flat, loss_of_thrust}, others specifically try to avoid transportation infrastructures~\cite{parachute_from_database}. One of the objectives of this paper is to propose unified objectives and requirements for EL algorithms, grounded on extensive safety considerations.

%\subsubsection{Other approaches to EL} To finish this literature review, it is also worth mentioning approaches considering the emergency control of the UAV for landing. In~\cite{external_conditions_EL}, the LZS algorithm takes into account parameters related to trajectory generation, such as wind field and nearby obstacles, in order to select a reachable landing spot. The approach proposed in~\cite{parafoil} studies the control of UAV after a parafoil has been opened in an emergency situation. An method to land UAV experiencing loss of thrust is proposed in~\cite{loss_of_thrust}.

\section{Certifying UAV urban operations}\label{sec:sora}

In this section, the process proposed in the current version of the SORA~\cite{SORA} is applied to a specific use case to demonstrate the difficulty to certify UAV urban operations. 

\subsection{The \droso~case study}\label{sec:medidelivery}

The use case considered in this paper is called \droso. It consists in using a UAV to deliver small automatic emergency equipment (e.g., defibrillator). It aims to shorten the intervention delay in cities. The UAV evolves in a complex airspace that may contain static obstacles (buildings, power lines, etc) and moving obstacles (other aircrafts).

\droso~is a rotary wing UAV with a span of around one meter. It flies at a height around 120 meters, leading to a typical ballistic vertical speed of $48.5m.s^{-1}$. Combining this with a Maximum Take-Off Weight (MTOW) of $7kg$ yields a kinetic energy of $8.23KJ$. \droso~flies above populated areas and goes Beyond Visual Line Of Sight (BVLOS).
%Finally, the \droso~UAV is equipped with a barometer, a GPS, an IMU, a FPV camera and a NADIR camera. This way, it can easily acquire data about its altitude, location, attitude, angular speed, as well as visual information. 
\subsection{Preliminary hazard analysis}\label{sec:nasa_hazards}

In~\cite{pha_drone_nasa}, Belcastro et al. proposed a comprehensive hazard analysis for generic civilian UAV operations. They study past accidents and conduct hazard analysis on various use cases to come up with fourteen main hazard categories such as loss of control, fly-away, lost communication, etc. The potential outcomes of these hazards can be divided into the two main risk categories proposed in the SORA: ground risks (GRC) and air risks (ARC). In this section, the hazard analysis from~\cite{pha_drone_nasa} is extended by conducting a severity analysis of hazardous outcomes related to ground risk. For each of the potential hazardous outcomes, we define severity levels (Table~\ref{tab:severity}). Due to space limitations the whole analysis is not presented, but the different outcomes and their associated severity are reported in Table~\ref{tab:hazardous}.

\begin{table}[ht]
    \centering
    \rowcolors{1}{}{lightgray}
    \begin{tabular}{>{\centering\arraybackslash}m{0.05\textwidth}|m{0.39\textwidth}}
        \rowcolor{gray} Rating & \multicolumn{1}{c}{Description} \\ \hline
        1 & \textbf{Negligible} - No effect \\
        2 & \textbf{Minor} - Slight injury or damage to the drone \\ 
        3 & \textbf{Serious} - Important injury or damage to critical infrastructures, environment \\ 
        4 & \textbf{Major} - Single fatal injury \\ 
        5 & \textbf{Catastrophic} - Multiple fatal injuries
    \end{tabular}
    \caption{Severity table}
    \label{tab:severity}
\end{table}

\begin{table}[ht]
    \centering
    \rowcolors{1}{}{lightgray}
    \begin{tabular}{>{\centering\arraybackslash}m{0.03\textwidth}|m{0.335\textwidth}|>{\centering\arraybackslash}m{0.045\textwidth}}
        \rowcolor{gray} ID & \multicolumn{1}{c|}{Hazardous outcomes} & Severity \\ \hline
        R1 & UAV causes accident involving ground vehicles & 5 \\
        R2 & UAV injures people on ground & 4 \\
        R3 & Post-crash fire that threatens wildlife and environment & 3 \\
        R4 & UAV collides with infrastructure (Building, bridge, power lines / sub-station, etc.) & 3 \\
        R5 & UAV crashes into parked ground vehicle & 2
    \end{tabular}
    \caption{Main ground risks}
    \vspace{-15pt}
    \label{tab:hazardous}
\end{table}

\subsection{Intended safety architecture}\label{sec:droso_safety_archi}
\begin{figure}[ht]
     \centering
     \includegraphics[width=0.48\textwidth]{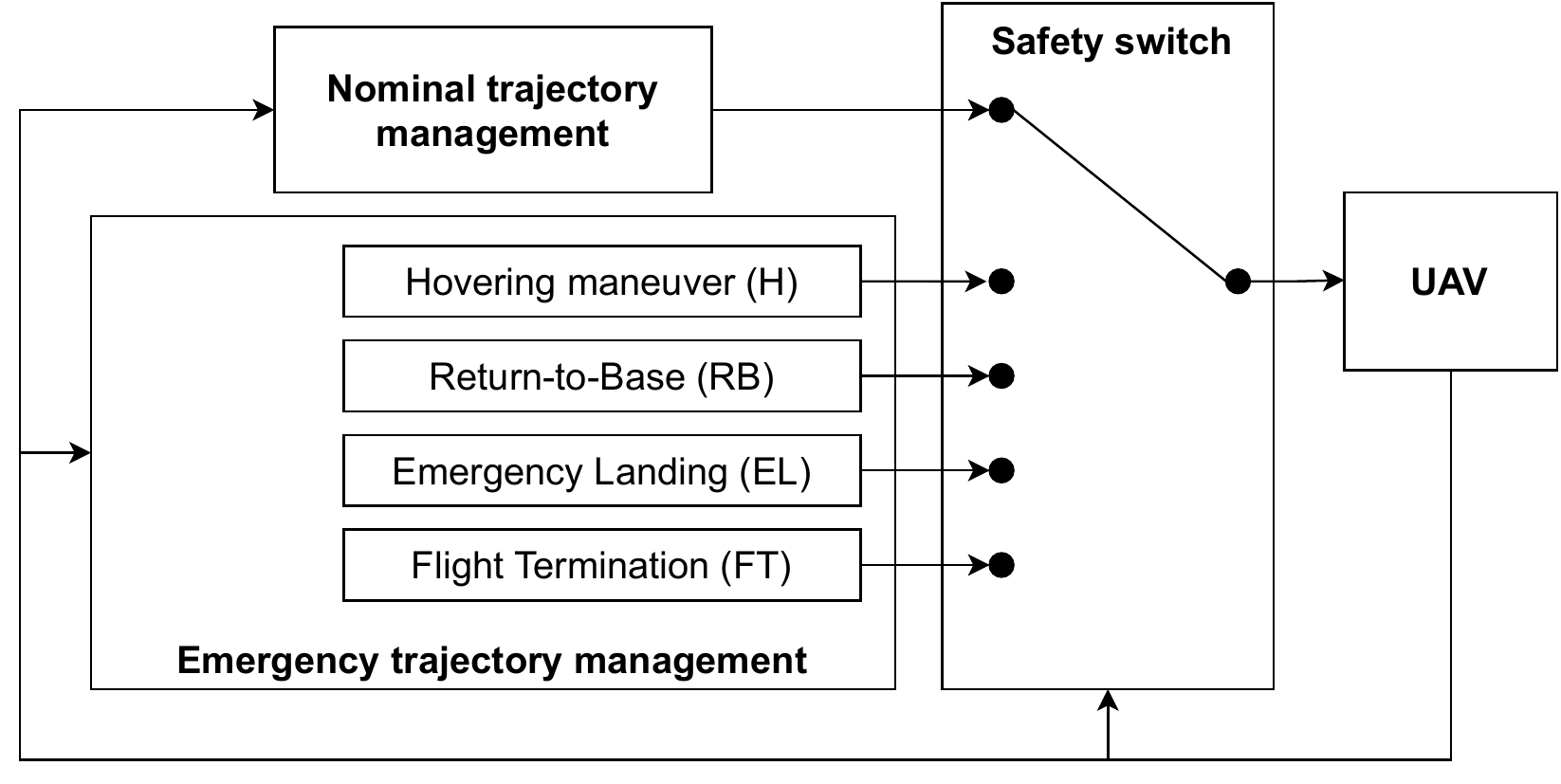}
     \caption{Architecture of internal functions \vspace{-10pt}}
     \label{fig:architecture_overview}
\end{figure}

Following the hazard analysis, an architecture of the internal functions of the UAV is proposed in Figure~\ref{fig:architecture_overview}. It has been designed as a continuous monitoring loop analyzing acquisition data to trigger the suitable emergency procedure when a critical anomaly is detected. The safety strategy can be described in more details as follows:
\begin{itemize}
    \item If the UAV faces a temporary unavailability of external services, an Hovering maneuver (H) is applied.
    \item If the UAV faces a permanent unavailability of communication services or on-board failures still allowing proper navigability, a Return-to-Base (RB) maneuver is applied.
    \item If the UAV faces a loss of navigation capabilities still allowing proper trajectory control (mainly localization and communication loss), an autonomous Emergency Landing maneuver (EL) is applied.
    \item If the UAV cannot ensure flight continuation or safe EL, then a Flight Termination maneuver (FT) is applied. In practice, it stops the engines and opens a parachute.
\end{itemize}

\subsection{Application of the SORA}

We apply the SORA to \droso~in order to evaluate the risks (GRC and ARC) and to identify the required safety counter-measures (see Section~\ref{sec:sora_overview}). %In the SORA, two risk classes are considered: the Ground Risk Class (GRC) evaluates the risk of collision with population or critical infrastructures and the Air Risk Class (ARC) evaluates the risk of mid-air collision with an inhabited aircraft. As explained earlier, the safety analysis proposed by the SORA consists in determining the GRC and ARC associated with a given UAV operation, as well as to identify the required strategies to mitigate the risk and ensure safe operations. 

\subsubsection{Initial risks identification}   
The determination of the GRC is based on the span, the typical kinetic collision energy of the UAV and the population density of the evolution and emergency areas. Based on the elements presented in Section~\ref{sec:medidelivery}, the resulting intrinsic GRC is 6. %The mid-air hazardous situations are considered by determining the ARC. 
The determination of the ARC is based on the characteristics of the aeronautic space. Since the maximum flight level is below 500 ft in a populated area, the resulting initial ARC is ARC-c.

\subsubsection{Mitigation}
We consider that \droso~is evolving within a dedicated corridor segregated from other UAV or manned aircraft airspace. Thus the mid-air collision avoidance is tied to the assurance of remaining inside this predefined corridor. In the remaining of this paper we do not consider specific Detect and Avoid mechanisms, hence the final ARC remains ARC-c.

Regarding ground risk, among the emergency procedures presented in Section~\ref{sec:droso_safety_archi} (Figure~\ref{fig:architecture_overview}), H and RB are degraded control modes that should not lead to ground collision, and cannot be used as GRC mitigation strategies. The remaining of this section discusses the possibility to use one of the three predefined GRC mitigation types (see Section~\ref{sec:sora_overview}) within \droso.

\begin{description}
    \item[M1 - Strategic mitigation] For urban UAVs like \droso, M1 mitigation, based on a low density ground buffer throughout the drone trajectory, cannot apply as the operation is not likely to be conducted entirely in low density area (busy roads, crowded areas). 
    \item[M2 - Reduction of ground impact effects] If we consider the hazardous situations presented in Table~\ref{tab:severity}, this strategy can reduce widely the risk to injure people directly (risk R2), decreasing the severity of an impact from 4 to 2. However, applying such M2 mitigation will not reduce the impact of the most severe outcomes, since a landing on a busy road (risk R1) could still cause fatal accidents. Hence, M2 cannot be considered sufficient to decrease the GRC for \droso. 
    \item[M3 - Emergency Response Plan (ERP)] An ERP can be designed for the \droso~use case, however, to be able to decrease the GRC the ERP needs to \textit{significantly reduce the number of people at risk}. As the major risk is almost immediate in case of an unwanted landing on a busy road, it is not possible to argue such a GRC lowering.
\end{description}

Hence, none of the mitigation solutions proposed in the current version of the SORA applies directly to \droso. The final GRC is at least 6 (7 if no M3 with medium robustness is proposed).

\subsubsection{Final safety objectives for \droso}
By combining the residual ground and air risk classes, the final SAIL allocated to \droso~is 5 (6 if no M3 is proposed). Knowing that the SAIL ranges from 1 to 7, one can consider that \droso~is a \textit{high risk} operation among the \textit{specific} category. Thus all the OSOs are requested and most of them at a high level of integrity and assurance. As is, the safety objectives require a high integrity and availability of the overall system from the drone to the pilot. Demonstrating a conformance to these objectives may lead to prohibitive validation and verification costs. This illustration of the SORA application motivates the need to propose alternative mitigation means such as EL along with a rigorous demonstration of its risk reduction and assurance enabling the applicant to alleviate the objectives over the whole UAV.

\section{Emergency Landing as a SORA mitigation}\label{sec:el_mitigation}

%\subsection{Overview}
%As shown in Section~\ref{sec:sora}, EL is not among the predefined mitigation strategies proposed in the SORA. 
The objective of this section is to introduce safety requirements for EL, so that it could be used straightforwardly as a SORA mitigation. In the SORA, a mitigation mean comes with a \textit{robustness} level, which is actually the combination of the severity reduction induced by the mitigation (called \textit{integrity} in SORA) and the confidence in this reduction (called \textit{assurance} in SORA).
Hence to be able to include EL in the SORA, we need to define in which SORA mitigation category EL is falling, and which requirements can be expressed to determine \textit{integrity} and \textit{assurance} levels. %This way, we can ensure that proposed EL implementations are designed effectively to reduce risks and can be trusted through a proper validation. 

When a crash is imminent, the mitigation can lower the collision risk as follows: M1 mitigation reduces the number of people at risk by ensuring that the drone will remain in a sparsely populated area, M2 mitigation limits the outcomes of the collision, and M3 mitigation reduces the people at risk by signaling the crash (e.g., evacuation of the area). Among the three mitigation categories, EL is closest to M1 as it intends to reduce the number of people at risks by identifying a safe landing zone. However, the SORA does not consider that safe landing zones can be actively identified from live data. Hence, we propose an adaptation of the M1 requirements for \textit{active-M1} mitigation techniques like EL. %To demonstrate consistency with the SORA, the proposed requirements are compared to M1 throughout this section. %Finally, we do not claim that the proposed requirement list for EL is  exhaustive, but we hope that it can serve as a good basis for the community, and that it can be complemented by future research.

\subsection{SORA integrity level for EL}\label{sec:integrity}

\begin{table*}[ht]
\caption{Level of Integrity Assessment Criteria for Emergency Landing}
\centering
\begin{minipage}{0.95\textwidth}
\centering
\rowcolors{1}{}{lightgray}
\begin{tabular}{m{0.1\textwidth}m{0.42\textwidth}|m{0.42\textwidth}}
\rowcolor{gray} \multicolumn{1}{c|}{Level} & \multicolumn{1}{c|}{Existing SORA criteria for M1 (Annex B of~\cite{SORA})} & \multicolumn{1}{c}{Proposed new criteria for EL (active-M1)}\\ \hline
\multicolumn{1}{c|}{Low} & \vspace{4pt} \begin{enumerate}[leftmargin=8pt]
\item A ground risk buffer with at least a 1 to 1 rule.
\item The applicant evaluates the area of operations by means of on-site inspections/appraisals to justify lowering the density of people at risk.
\end{enumerate} ~\vspace{-8pt} & \vspace{4pt} \begin{enumerate}[leftmargin=8pt]
\item The selected landing zones do not contain high risk areas\footnote{Certain high risk areas (e.g., occupied by people) can be used if a provably efficient mitigation is in place for the hazards involved (e.g., parachute).} (As defined in Table~\ref{tab:severity}).
\item The method is effective under the conditions of the operation (specific city, flight altitude, time of the day, season, etc.).\end{enumerate} ~\vspace{-8pt}\\ 
\multicolumn{1}{c|}{Medium} & 
\vspace{4pt} \begin{enumerate}[leftmargin=8pt]
\item Ground risk buffer takes into account: \begin{itemize}[leftmargin=12pt]
\item Improbable single malfunctions or failures,
\item Meteorological conditions (e.g., wind),
\item UAV latencies, behavior and performance.
\item UAV behavior when activating measure,
\item UAV performance.
\end{itemize} 
\item The applicant uses authoritative density data relevant for the area and time of operation.
\end{enumerate}~\vspace{-8pt} & 
\vspace{4pt} \begin{enumerate}[leftmargin=8pt]
\item Landing zone selection takes into account\footnote{Selected landing zone should be far enough from hazardous areas to guarantee that adverse conditions will not lead the UAV to hazardous situations.}: 
\begin{itemize}[leftmargin=12pt]
%\item UAV loss of maneuverability (degraded control modes),
%\item Meteorological conditions (wind, rain),
%\item UAV performance.
\item Improbable single malfunctions or failures,
\item Meteorological conditions (e.g., wind),
\item UAV latencies, behavior and performance.
\item UAV behavior when activating measure,
\item UAV performance.
\end{itemize}
\end{enumerate} 
~\vspace{-4pt} 
\\ 
\multicolumn{1}{c|}{High} & \vspace{4pt} Same as Medium \vspace{4pt} & \vspace{4pt} Same as Medium \vspace{4pt} \\ 
\end{tabular}
\end{minipage}
\label{tab:integrity}
\end{table*}

In Table~\ref{tab:hazardous}, we have identified that the most severe potential outcome of a UAV urban landing is to land on a busy road. Indeed, for all types of landing (parachute, controlled landing, crash), a UAV reaching the ground on a busy road can always result in an accident with multiple fatalities. 
The second outcome that can cause death is when the UAV lands on people. However, this risk can be mitigated using an effective M2 mitigation to reduce the effects of the impact. 

Hence, to claim any severity reduction, i.e., a low integrity in the SORA parlance, an EL module should avoid roads at all costs, and should prevent the UAV from landing on populated areas without an efficient mechanism to reduce the kinetic energy of the UAV.
Additionally to claim a minimal integrity, the geometry of the selected landing zone should take into account the conditions of operation that may influence the deviation during the landing maneuver (potentially performed by a parachute). 
For example, if the UAV lands with parachute opened at a given altitude, the buffer from roads must take into account the typical parachute drift in nominal conditions. 

As proposed for the initial M1, any higher integrity level can be achieved by taking into account adversary conditions and failures in the landing zone definition.
The proposed integrity requirements for EL are summarized in Table~\ref{tab:integrity}. For comparison, the corresponding criteria for M1 mitigation are also reported.

\subsection{SORA assurance level for EL}\label{sec:assurance}

\begin{table*}[ht]
\caption{Level of Assurance Assessment Criteria for Emergency Landing}
\centering
\begin{minipage}{0.95\textwidth}
\centering
\rowcolors{1}{}{lightgray}
\begin{tabular}{m{0.1\textwidth}m{0.42\textwidth}|m{0.42\textwidth}}
\rowcolor{gray} \multicolumn{1}{c|}{Level} & \multicolumn{1}{c|}{Existing SORA criteria for M1 (Annex B of~\cite{SORA})} & \multicolumn{1}{c}{Proposed new criteria for EL (active-M1)}\\ \hline
\multicolumn{1}{c|}{Low} & \vspace{4pt} \begin{enumerate}[leftmargin=8pt]
    \item The applicant declares that the required level of integrity is achieved. \end{enumerate} ~\vspace{-8pt} & \vspace{4pt} \begin{enumerate}[leftmargin=8pt]
        \item The applicant declares that the required level of integrity is achieved. \end{enumerate} ~\vspace{-8pt} \\ 
\multicolumn{1}{c|}{Medium} & 
\vspace{4pt} \begin{enumerate}[leftmargin=8pt]
\item Supporting evidence to claim the required level of integrity has been achieved (testing, analysis, simulation, inspection, design review, experience).
\item The density data used is an average density map for the date/time of the operation from a static sourcing (verified by applicable authority).
\end{enumerate}~\vspace{-4pt} & 
\begin{enumerate}[leftmargin=8pt] 
\item Supporting evidence to claim the required level of integrity has been achieved (testing on public datasets, testing in context\footnote{To ensure safety, real world tests can be realized by conducting pre-flights with a small UAV, containing the same camera as the real system.}).
\item The video data used for in-context testing are recorded and verified by applicable authority.
\item Safety monitoring techniques are in place to ensure proper behavior of any function relying on complex computer vision or machine learning.
\end{enumerate} ~\vspace{-8pt}\\ 
\multicolumn{1}{c|}{High} & \vspace{4pt} \begin{enumerate}[leftmargin=8pt] 
\item The claimed level of integrity is validated by a competent third party.
\item The density data used is a near-real time density map from a dynamic sourcing and applicable for the date/time of the operation.
\end{enumerate} ~\vspace{-8pt} & \vspace{4pt} \begin{enumerate}[leftmargin=8pt]
%\item A procedure for continuous validation is implemented to collect data during mission and improve EL assurance as new missions are executed.
\item The claimed level of integrity is validated by a competent third party. 
\item The method was extensively validated under a wide range of external conditions (lighting, weather).\end{enumerate} ~\vspace{-4pt} \\ 
\end{tabular}
\end{minipage}
\label{tab:assurance}
\end{table*}

The assurance requirement of the Table~\ref{tab:assurance} explains how the confidence in a mitigation is commensurate to the conducted validation and verification activities. 

As for M1, the minimal level of assurance is achieved by a simple declaration of the applicant. 
As shown in Section~\ref{sec:literatureEL}, LZS relies on complex computer vision functions, on which it is difficult to provide a rigorous specification of the expected behavior of the learned model. 
This way, traditional safety assurance practices conducted during the design process are not sufficient to ensure safe behavior of the EL system~\cite{ml_safety}. 
Hence to claim a higher level of assurance, the applicant must justify that safety monitoring techniques checking the outputs of the LZS model at runtime has been designed. 
Moreover the applicant must guarantee that both the main system and the monitor are working as expected, tests must be conducted in the specific context of the operation. 
The applicant should also validate his approach under different external conditions in order to know the validity domain of the proposed algorithm. Finally, for high assurance, all tests should be designed and validated jointly with a competent third party. %The proposed assurance requirements for EL are summarized in Table~\ref{tab:assurance}.

\section{EL implementation proposal}\label{sec:implementation}

To comply with the EL integrity requirements from Table~\ref{tab:integrity}, a possible two step implementation of EL is:
\begin{enumerate}
    \item Select an area far from busy roads.
    \item Go to this area and open a parachute.
\end{enumerate}
Indeed, from what was seen above (Table~\ref{tab:severity}), if this algorithm is implemented successfully with good assurance, no fatality should occur. In this section, we focus on the landing zone selection module, and a Busy Road Detection (BRD) system is presented. To comply with the assurance requirements of the Table \ref{tab:assurance}, we also propose a new approach to monitor the behavior of the BRD model at runtime in order to increase confidence in its predictions, as well as the associated safety architecture for integration. %Such monitoring methods are necessary to guarantee \textit{Medium} assurance level for the EL module (Table~\ref{tab:assurance}).

\begin{figure*}[ht]
    \centering
    \includegraphics[width=0.85\textwidth]{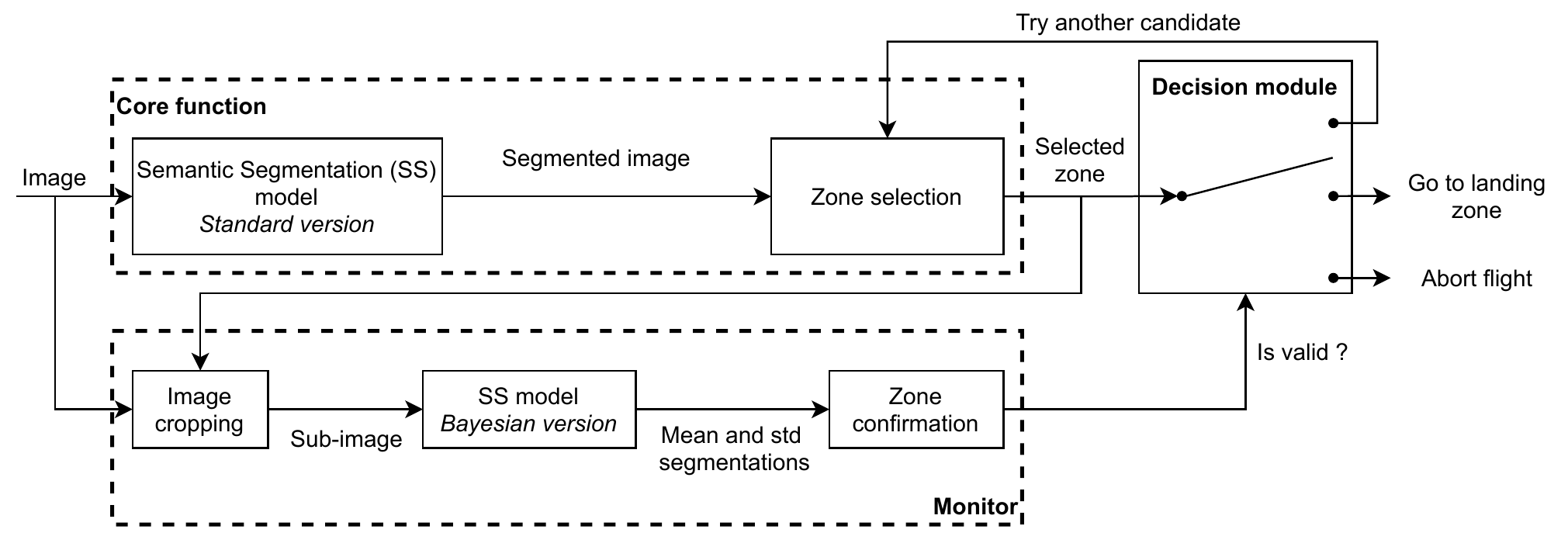}
    \caption{Safety architecture for the proposed Landing Zone Selection module. \vspace{-10pt}}
    \label{fig:architecture}
\end{figure*}

\subsection{Landing zone selection architecture}\label{sec:architecture}

The general architecture proposed for landing zone selection is shown in Figure~\ref{fig:architecture} is derived from the typical Computer/Monitor safety pattern. It is composed of the core function, which selects a landing zone candidate from an input image, and the monitor, verifying that this candidate is safe. Then, the Decision Module (DM) is in charge of deciding what actions to take. If the monitor confirms the proposed zone, then the DM will trigger landing execution. If the zone is rejected by the monitor, the DM will either request a new trial or abort the flight if an additional trial cannot be safely performed. %, depending on the criticality of the situation.
The core function intends to address Low and Medium Integrity requirements (Table~\ref{tab:integrity}), whereas the monitor deals with the Medium-3 assurance requirement (Table~\ref{tab:assurance}).

\subsection{Implementation details}

\begin{figure}[ht]
    \centering
    \begin{subfigure}{0.4\textwidth}
    \includegraphics[width=\textwidth]{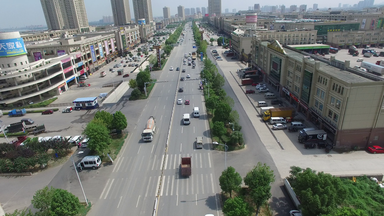}
    \caption{Image from the training set.}\label{fig:example_semseg_im}
    \end{subfigure}
    
    \begin{subfigure}{0.4\textwidth}
    \includegraphics[width=\textwidth]{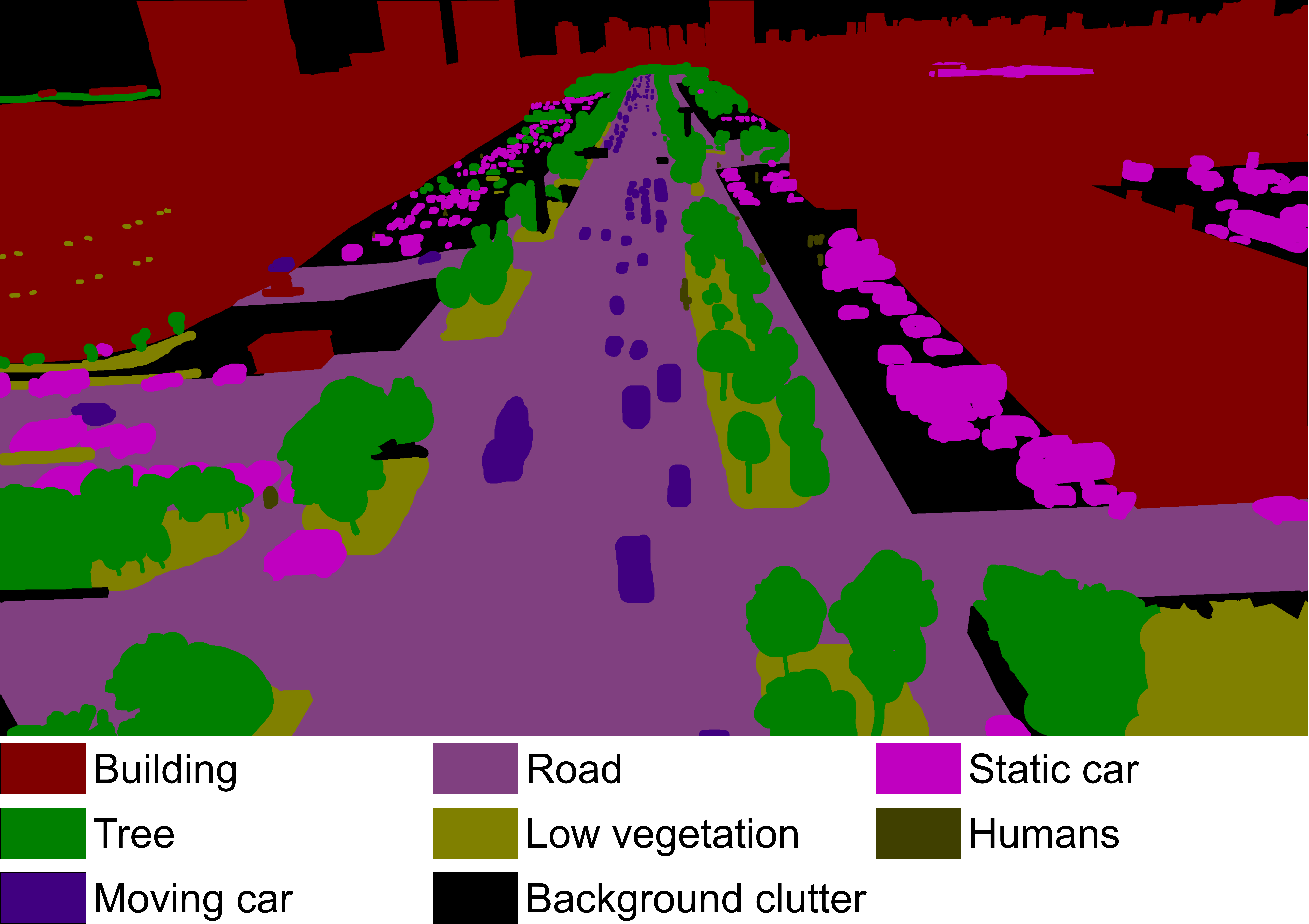}
    \caption{Corresponding labels.}\label{fig:example_semseg_lab}
    \end{subfigure}
    
    \caption{Example image and corresponding labels from the UAVid dataset~\cite{uavid}. \vspace{-15pt}}
    \label{fig:example_semseg}
\end{figure}

\begin{figure*}[ht]
    \centering
    \begin{subfigure}{0.87\textwidth}
    \includegraphics[width=\textwidth]{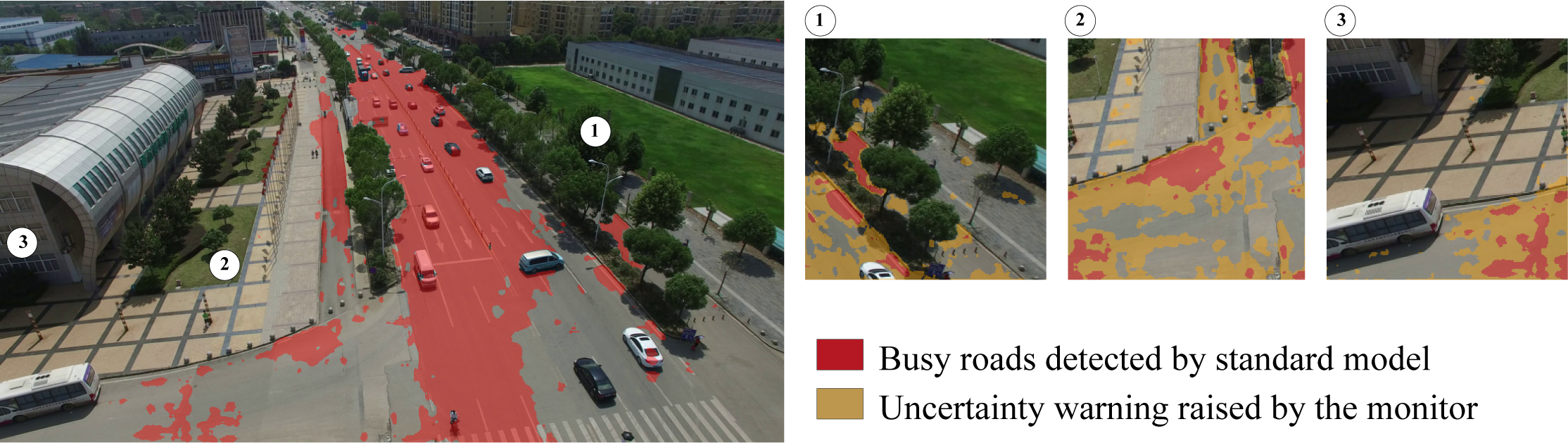}
    \caption{Image from the UAVid test set.}\label{fig:results_indist}
    \end{subfigure}
    
    \begin{subfigure}{0.87\textwidth}
    \includegraphics[width=\textwidth]{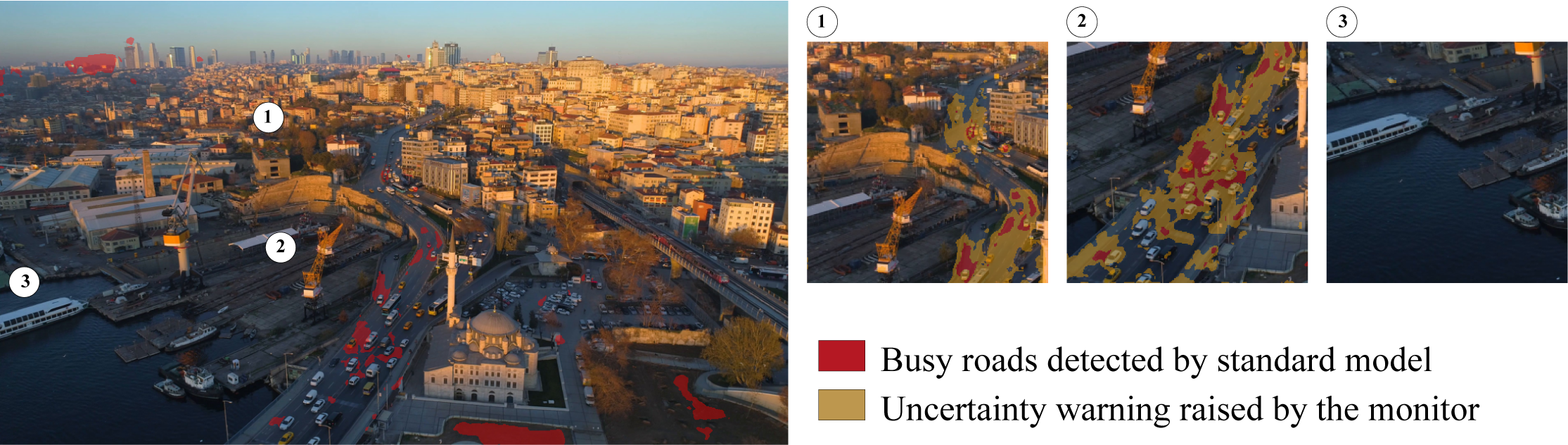}
    \caption{Image from outside the UAVid dataset.}\label{fig:results_ood}
    \end{subfigure}
    
    \caption{Semantic segmentation and associated monitoring results on two example images. \vspace{-10pt}}
    \label{fig:results}
\end{figure*}

The core function is based on semantic segmentation, which consists in assigning a label from a set of predefined categories to each pixel of an image~\cite{survey_semseg}. In particular, here we aim at building a semantic segmentation model able to determine if a given pixel is part of a busy road or not. To do this, we use a model called Multi-Scale-Dilation net (MSDnet), proposed in~\cite{uavid}. It was trained on the UAVid dataset, which is composed of 300 high-resolution images captured from a UAV, in slanted views. UAVid images are densely labeled for semantic segmentation with 8 classes (Figure~\ref{fig:example_semseg}). %(building, road, static car, tree, low vegetation, human, moving car and background clutter). 
From this dataset, we cannot explicitly detect busy roads, instead we consider that all pixel corresponding to \textit{road} and \textit{cars} should be avoided. Then, for a given on-board image, the EL system can use the predicted segmentation to identify areas far from busy roads, to which the UAV can navigate and terminate the flight by opening its parachute. However, as most successful semantic segmentation models, MSDnet was trained using machine learning, and it is hard to obtain strong guarantees about its robustness. For instance, it tends to make mistakes on out-of-distribution data, as illustrated by the left image in Figure~\ref{fig:results_ood}. This image was extracted from a drone footage that does not belong to the UAVid dataset, and which presents challenging lighting conditions as it was taken at sunset. To address this issue, we propose to monitor the predictions using the model's self uncertainty.

However, like most deep learning models, MSDnet produces point-estimates as outputs and knowledge about the confidence of the model in its predictions is unknown~\cite{bayesian_NN}. In other words, for an image $I$, a given pixel $I_{ij}$ and a given category $y_k$, the score $s_{ijk}$ obtained after the softmax layer of the model does not represent the confidence of the network that $I_{ij}$ belongs to $y_k$. Basically, from a standard neural network, the only relevant information that can be extracted from the final softmax scores is the predicted class $y$:
\begin{equation}\label{eq:standard}
    y = y_k \iff s_{ijk} > s_{ijk'}, \forall k' \neq k
\end{equation}
A possible solution to compute uncertainty in the predictions of MSDnet is to use a Bayesian version of the same model~\cite{bayesian_semseg}. In practice, we use the Monte Carlo dropout method to convert the standard MSDnet model into a Bayesian one (BMSDnet)~\cite{bayesian_semseg}. It consists in maintaining the dropout layers active at inference time\footnote{In practice, we use a dropout rate of 0.5 for all relevant MSDnet layers}, which can be shown to be mathematically equivalent to an  approximation to the probabilistic deep Gaussian process~\cite{dropout_bayesian}. To obtain uncertainty from this simple Bayesian MSDnet, the input image must be processed several times. As different neurons are dropped each time, each inference results in different predictions. Then, for each pixel $I_{ij}$ and each category $y_k$, we can compute both the empirical mean $\mu_{ijk}$ and standard deviation $\sigma_{ijk}$ of the different values of $s_{ijk}$ predicted by BMSDnet. The standard deviation $\sigma_{ijk}$ is a good proxy for MSDnet uncertainty regarding the fact that $I_{ij}$ belongs to class $y_k$. The intuition behind this approach is that a neural network trained with dropouts has redundant connections for its highly certain predictions, making them less impacted by the removal of some neurons.

Finally, EL is intended to serve as a critical safety function, thus classifying a busy road to another category can lead to catastrophic outcomes. Hence, the monitor is designed to be conservative by over-approximating the road category. To do so, once we have access to the probability distribution for a given pixel, the 99.7\% confidence interval of the predictions can be computed and tested against a small decision threshold $\tau$. Hence, $I_{ij}$ is considered safe (i.e., it does not belong to the road category) if and only if it verifies
\begin{equation}\label{eq:monitor}
    \mu_{ij} + 3\times\sigma_{ij} \leq \tau.
\end{equation}
We note that in Equation~\ref{eq:monitor}, the index related to category was dropped as we only consider the \textit{busy road} category. As UAVid is composed of 8 categories, we choose $\tau=0.125$ in order to make sure that the \textit{road score} is lower than a random guess. In practice, Equation~\ref{eq:monitor} must be verified for the three UAVid categories that make up the busy road category. %Although both Equation~\ref{eq:standard} and \ref{eq:monitor} produce a segmentation, they represent very different information about the original image. The former represents the raw model prediction, while the latter only accepts pixels with high certainty of not being a road.

In practice, running Bayesian inference with MSDnet on such large images (3840x2160) is prohibitively slow as it requires to run the model several times. Indeed, EL is a critical function that must run fast in case of an emergency. In addition, the embedded computing power for medium size UAVs is often limited. This important constraint justify the selected monitoring architecture presented in Figure~\ref{fig:architecture}, where a candidate landing zones is pre-selected using the standard MSDnet, and only smaller sub-images are passed to the monitor in order to verify the proposals before triggering landing execution. Our first experiments using a Nvidia Quadro P5000 GPU showed that if prediction statistics are computed on 10 samples, the monitor verifies a 1024x1024 image in less than 5 seconds, whereas it takes over a minute for the full image.

\subsection{First results}\label{sec:results}

Preliminary experiments were conducted to evaluate qualitatively the proposed approach for road detection and runtime monitoring. Two example images can be seen in Figures~\ref{fig:results}. In Figure~\ref{fig:results_indist}, an image from the test set of UAVid was segmented using MSDnet (left), this image should be part of the test set to comply with requirements Medium-1, from the Assurance table (Table~\ref{tab:assurance}). Then, three sub-images containing areas with and without roads were selected manually to illustrate the behavior of the monitor. They are shown on the right and their top left corners are localized on the main image for ease of readability. The second image tested (Figure~\ref{fig:results_ood} was collected online and does not belong to UAVid, it could be used for proving the High-2 Assurance item. It represents very harsh conditions for our approach as the altitude of the drone is different from UAVid images and the image was taken at sunset, involving complex lighting conditions.

The proposed MSDnet seems to perform reasonably well on the image from UAVid, but it clearly fails on the out-of-distribution image. However, the monitor seems to be able to trigger an uncertainty warnings for a large part of the road areas that was not covered by the core model. We also note that no warning is raised for Figure~\ref{fig:results_ood}-3, as expected.

\section{Conclusion}\label{sec:conclusion}
In this work, we demonstrated through a case study that it is currently very hard to certify UAV operations in densely populated areas due to the lack of acceptable mitigation means in the SORA. Emergency landing, which selects landing zones actively using data from on-board cameras, appears to be a promising research direction to help solve this problem. Hence, a second contribution was to define a set of requirements consistent with SORA, so that EL could be considered as an applicable mitigation mean. Finally, we proposed an implementation of landing zone selection based on semantic segmentation, as well as a safety architecture to monitor machine learning components at runtime to comply with the proposed requirements. The first results illustrate the monitor capability to discard large road areas unseen by the model.

In future work, the preliminary results presented here should be complemented by a formal quantitative study to ensure compliance with the identified Assurance requirements. In addition, many regions containing roads are missed by the monitor. Hence, it will be worth investigating other segmentation models, including lightweight ones in order to be able to run on on-board GPUs. Training on other datasets and using other uncertainty estimation techniques could also be investigated in future research. Finally, hybrid methods combining learning-based techniques with using public databases could be envisioned to improve emergency landing. 

\bibliographystyle{IEEEtran}
\bibliography{IEEEabrv,references.bib}

\end{document}